\documentclass[runningheads]{llncs}

\usepackage{graphicx}
\usepackage{comment}
\usepackage{color}
\usepackage{hyperref}
\usepackage{soul}
\usepackage{pdflscape}
\usepackage{algorithmic}
\usepackage[ruled,lined,linesnumbered,longend]{algorithm2e}
\usepackage{amssymb,amsfonts,textcomp,amsmath}

\begin{document}
\title{Compact Optimization Algorithms with Re-sampled Inheritance}
%
%
\author{Giovanni Iacca\inst{1}\orcidID{0000-0001-9723-1830}
\and 
Fabio Caraffini\inst{2}\orcidID{0000-0001-9199-7368}
}
%
%
\institute{
Department of Information Engineering and Computer Science,\\ University of Trento, Povo 38123, Italy\\
\email{giovanni.iacca@unitn.it}
\and
Institute of Artificial Intelligence, School of Computer Science and Informatics,\\ 
De Montfort University, Leicester LE1 9BH, UK\\
\email{fabio.caraffini@dmu.ac.uk}
}
\maketitle 
\begin{abstract}
Compact optimization algorithms are a class of Estimation of Distribution Algorithms (EDAs) characterized by extremely limited memory requirements (hence they are called ``compact"). As all EDAs, compact algorithms build and update a probabilistic model of the distribution of solutions within the search space, as opposed to population-based algorithms that instead make use of an explicit population of solutions. In addition to that, to keep their memory consumption low, compact algorithms purposely employ simple probabilistic models that can be described with a small number of parameters. Despite their simplicity, compact algorithms have shown good performances on a broad range of benchmark functions and real-world problems. However, compact algorithms also come with some drawbacks, i.e. they tend to premature convergence and show poorer performance on non-separable problems. To overcome these limitations, here we investigate a possible algorithmic scheme obtained by combining compact algorithms with a non-disruptive restart mechanism taken from the literature, named Re-Sampled Inheritance (RI). The resulting compact algorithms with RI are tested on the CEC 2014 benchmark functions. The numerical results show on the one hand that the use of RI consistently enhances the performances of compact algorithms, still keeping a limited usage of memory. On the other hand, our experiments show that among the tested algorithms, the best performance is obtained by compact Differential Evolution with RI.
\end{abstract}
\section{Introduction}\label{intro}
Compact Optimization \cite{compactOptimization} is a branch of Computational Intelligence Optimization devoted to the study of optimization algorithms characterized by limited memory requirements. From an algorithmic point of view, compact algorithms belong to the family of the Estimation of Distribution Algorithms (EDAs)~\cite{EDAbook}, i.e. algorithms that instead of evolving a population of solutions (as is typically done in population-based optimization algorithms, such as Evolutionary Algorithms and Swarm Intelligence algorithms), build and update a probabilistic model of the distribution of solutions within the search space. Depending on the specific probabilistic model (Gaussian, binomial, etc.), different EDAs can be implemented. In this regard, the specificity of compact algorithms is that they employ a separate distribution for each variable of the problem, and update it as long as the evolutionary process proceeds. Therefore, differently from population-based algorithms where at least $n$ $D$-dimensional arrays need to be stored in memory (being $n$ the population size and $D$ the problem dimension), compact algorithms need to store only a much more compact ``Probability Vector'' ($\textbf{PV}$) that describes the parameters of the probabilistic model. For instance, binary-encoded compact algorithms use as \textbf{PV} a single $D$-dimensional array $\textbf{p} = [p_1, p2, \dots, p_D]$. Each $p_i \in [0,1]$, $i = 1, 2, \dots, D$, represents the probability that the $i-th$ variable has value 1 (i.e., the relative frequency in a corresponding ``virtual population" of $N_p$ individuals, with $N_p$ a parameter of the algorithm). Similarly, real-valued compact algorithms based on Gaussian distributions use as $\textbf{PV}$ two $D$-dimensional arrays: an array of means $\boldsymbol{\mu} = [\mu_1, \mu_2, \dots, \mu_n]$ and an array of variances $\boldsymbol{\sigma}= [\sigma_1, \sigma_2, \dots, \sigma_n]$, describing a (normalized) Gaussian distribution of each variable in the search space.

In the past two decades, the Compact Optimization concept has been declined in a number of compact algorithms, sparkling from the seminal works by Harik et al. \cite{cGA} and Corno et al. \cite{SG}, who devised a similar algorithm dubbing it respectively ``compact Genetic Algorithm'' (cGA) and ``Selfish Gene'' (SG). The family of compact algorithms was then extended to include improved versions of cGA \cite{cGAElitism,familycGA}), real-valued cGA (rcGA) \cite{rcGA}, compact Differential Evolution (cDE) \cite{cDE} and many of its variants \cite{globalcDE,superfitcDE,cDELight,coDE,obcDE,nacDE,CMcDE,ensemblecDE,McDE,disturbedcDE}, compact Particle Swarm Optimization (cPSO) \cite{cPSO}, compact Bacterial Optimization (cBFO) \cite{cBFO}, and, more recently, compact Teaching-Learning Based Optimization (cTLBO) \cite{cTLBO,rcTLBO}, compact Artificial Bee Colony (cABC) \cite{EcABC,cABC}, and compact Flower Pollination Algorithm (cFPA) \cite{cFPA}.

Due to their limited usage of memory, compact algorithms are particularly suited for embedded devices, such as Wireless Sensor Networks motes, wearable devices, embedded controllers for robots and industrial plants, etc. Unsurprisingly, the literature abounds with successful examples of compact algorithms applications based on this kind of devices: for instance, cDE has been applied especially in control applications, such as real-time hardware-in-the-loop optimization of a control system for a permanent-magnet tubular linear synchronous motor \cite{cDE}, or real-time trajectory optimization of robotic arms \cite{cDELightRobot,disturbedcDE} and Cartesian robots \cite{McDE}. In \cite{familycGA}, cGA was applied to micro-controller design, while cPSO was used for optimizing a power plant controller in \cite{cPSO}. In \cite{cABC-WSN}, cABC was used for topology control in Wireless Sensor Networks. The more recent cTLBO was instead applied to train Artificial Neural Networks in \cite{rcTLBO}.

In this paper, we aim to push forward this research area by tackling the two main drawbacks of compact algorithms, i.e. their tendency to premature convergence (as they do not keep an actual population, they do not maintain explicitly diversity), and a poorer performance on non-separable problems (which is due to the fact that they process each variable separately). To overcome these limitations, we study the effect of a special case of restart named Re-Sampled Inheritance (RI) \cite{RICaraffini2013a,RICaraffini2013b}, which simply generates a random solution and then recombines it -by using an exponential crossover operator similarly to Differential Evolution- with the best solution detected so far by the compact algorithm. Previous works \cite{LEGO2018RICMAES,RICaraffini2013b} have shown that the RI mechanism is a simple yet effective way to improve the performance of an optimization algorithm, as it allows to escape from local optima while preserving some of the information from the current best, thus guaranteeing a kind of inheritance and avoiding an excessively disruptive restart (compared to other restart mechanisms). Our hypothesis is then that on the one hand the re-sampling should allow compact algorithms to escape from local optima, on the other hand the inheritance mechanism -since it processes blocks of variables- should enable an overall performance improvement especially (but not only) on non-separable problems. Moreover, as the RI mechanism only needs to sample a random solution and recombine it with the current best, it does not require any additional memory with respect to a compact algorithm, therefore allowing to keep a low memory consumption. 

To assess the effect of RI on different compact algorithms, we apply it separately to four (real-valued) compact algorithms taken from the literature, namely cDE (specifically, its `light'' version \cite{cDELight}), rcGA, cPSO, and cBFO, and perform extensive tests on the CEC 2014 benchmark \cite{cec2014} in 10, 50 and 100 dimensions. 

The rest of this paper is organized as follows: Section \ref{background} presents the background concepts on Compact Optimization. Section \ref{algorithm} describes the general algorithmic scheme which combines compact algorithms with Re-Sampled Inheritance. The numerical results are then presented and discussed in Section \ref{results}. Finally, Section \ref{conclusions} concludes this work.

\section{Background}\label{background}
In the rest of this paper, we focus on real-valued compact algorithms as they have been shown to perform better than their binary-encoded counterparts \cite{rcGA}. In case of real values, the general structure of a compact algorithm is quite straightforward and can be described as follows. First of all, for each i-th variable a Gaussian Probability Distribution Function (PDF) is considered, truncated within the interval $\left[-1,1\right]$, with mean $\mu_i$ and standard deviation $\sigma_i$ taken from the Probability Vector $\textbf{PV}=[\boldsymbol{\mu},\boldsymbol{\sigma}]$. The height of the PDF is normalized in order to keep its area equal to $1$.

At the beginning of the optimization process, for each design variable $i$, $\mu_i=0$ (unless Re-Sampled Inheritance is used, see next section) and $\sigma_i=\lambda$, where $\lambda$ is a large positive constant (e.g. $\lambda=10$), such that it simulates a uniform distribution (thus exploring the search space). Subsequently, a starting individual, $\bf{elite}$, is generated by sampling each i-th variable from the corresponding PDF. For more details about the sampling mechanism, see \cite{rcGA}. 

Then, the iterative process starts. At each step, depending on the specific compact algorithm, a candidate solution $\bf{x}$ is generated by sampling one or more individuals from the current \textbf{PV}. E.g., in rcGA \cite{rcGA}, $\bf{x}$ is obtained by generating a single individual and recombining it with $\bf{elite}$ with binomial crossover. In cDE \cite{cDE}, $\bf{x}$ is obtained by generating a mutated individual (for instance sampling three individuals from $\textbf{PV}$ and applying the rand/1 DE mutation), and then recombining it with the current $\bf{elite}$ by using either binomial or exponential crossover. Other compact algorithm paradigms, such as cPSO and cBFO, use the same mechanism for sampling new individuals, but apply different algorithmic operators inspired from the corresponding biological metaphor to generate a new candidate solution $\bf{x}$. In all cases, the fitnesses of $\bf{elite}$ and $\bf{x}$ are compared and, according to the chosen elitism scheme (persistent or non-persistent, see \cite{compactOptimization}), $\bf{elite}$ is replaced by $\bf{x}$. Furthermore, the fitness comparison is used to update the $\textbf{PV}$, i.e. it changes its $\mu$ and $\sigma$ values by ``moving'' the Gaussian PDF towards the better solution and ``shrinking'' the PDF around it. Details for this update mechanism are given in \cite{rcGA}. This iterative process is executed until a certain stop condition is met. The pseudo-code of a general real-valued compact algorithm is given in Algorithm \ref{alg:generalCompactAlgorithm}.
\vspace{-0.5cm}
\begin{algorithm}[!ht]\caption{General structure of a compact algorithm}\label{alg:generalCompactAlgorithm}
    initialize $\bf{PV}=[\boldsymbol{\mu},\boldsymbol{\sigma}]$\;
    generate $\bf{elite}$ by means of $\bf{PV}$\;
    \While{stop condition is not met}{
    	generate candidate solution $\bf{x}$ (according to the specific operators)\;
    	compare fitness of $\bf{x}$ and $\bf{elite}$\;
    	\If{elite replacement condition is true}{
       		$\bf{elite}=\bf{x}$\;
    	}
        update $\bf{PV}$\;
    }
    \Return \textbf{elite}
\end{algorithm}
\vspace{-1cm}

\section{Compact Optimization Algorithms with Re-Sampled Inheritance}\label{algorithm}
The general scheme of a compact algorithm with Re-Sampled Inheritance is shown in Algorithm \ref{alg:compactAlgorithmWithRI}. The only difference w.r.t. the original compact algorithm shown in Algorithm \ref{alg:generalCompactAlgorithm} is the RI component, which enables the restart mechanism with inheritance, and is activated at the end of each execution of the compact algorithm (that is continued for a given \% of the total budget).

The Re-Sampled Inheritance (see \cite{RICaraffini2013a,RICaraffini2013b} for more details) first randomly generates a solution $\bf{x}$ from a uniform distribution within the given search space. Then, it recombines $\bf{x}$ with the current best solution $\mathbf{x_{best}}$ by applying the exponential crossover used in Differential Evolution. More specifically, a random initial index is selected in $[0,D)$, and the corresponding variable is copied from $\mathbf{x_{best}}$ into $\bf{x}$. Then, as long as a (uniform) random number $rand(0,1)$ is less than or equal to $Cr$, the design variables from $\mathbf{x_{best}}$ are copied into the corresponding positions of $\bf{x}$, starting from the initial index. $Cr$, the crossover rate, is a parameter affecting the number of variables inherited from $\bf{x_{best}}$, and is set as in \cite{cDELight}, i.e. $Cr=1/\sqrt[D\alpha]{2}$, where $D\alpha$ is the expected number of variables that are copied from $\mathbf{x_{best}}$. As soon as $rand(0,1) > Cr$, the copy process is interrupted. The copy is handled as in a cyclic buffer, i.e. when the D-th variable is reached during the copy process the next to be copied is the first one. When the copy stops, the fitness of $\bf{x}$ is compared with that of $\mathbf{x_{best}}$. If the newly generated solution $\bf{x}$ outperforms the current best solution, the latter gets updated (i.e. $\mathbf{x_{best}=\bf{x}}$). The compact algorithm is then restarted after setting its mean value $\boldsymbol{\mu}$ (that is used in the Probability Vector $\bf{PV}$) equal to the new restarted point, i.e. $\boldsymbol{\mu}=\bf{x}$. This way, the new initial distribution is centered in a new point which, despite being randomly sampled, still contains some inheritance from the current best solution. At the end of each compact optimization routine, see Algorithm \ref{alg:generalCompactAlgorithm}, an elite solution is returned and compared for replacement against the current best solution, as shown in Algorithm \ref{alg:compactAlgorithmWithRI}.
\vspace{-0.5cm}
\begin{algorithm}[!ht]\caption{Compact algorithm with Re-Sampled Inheritance} 	\label{alg:compactAlgorithmWithRI}
    generate a random solution $\bf{x}$ in the search space and  set $\mathbf{x_{best}}=\bf{x}$\;
	\While{stop condition is not met}{
    	// Compact algorithm{}
        \\
    	set $\boldsymbol{\mu}=\bf{x}$ and run compact algorithm as in Alg. \ref{alg:generalCompactAlgorithm} (for a \% of the budget)\;
        {}\
        \If{$\bf{elite}$ is better than $\mathbf{x_{best}}$}{
    		$\bf{x_{best}}=\bf{elite}$\;
    	}
        // Re-Sampled Inheritance
        \\
    	generate a random solution $\bf{x}$ (from a uniform distribution)\;
    	generate $i=round \left(D \cdot rand\left(0,1\right)\right)$\;
        $\bf{x}[i]=\bf{x_{best}}[i]$\;
        $k=1$\;
    	\While{$rand\left(0,1\right)\leq Cr$ \bf{and} $k<D$}{
    		$i=i+1$\;
    		\If{$i==D$}{
    			$i=1$\;
    		}
            $\bf{x}[i]=\bf{x_{best}}[i]$\;
            $k=k+1$\;
    	}
        \If{$\bf{x}$ is better than $\bf{x_{best}}$}{
    		$\bf{x_{best}} = \bf{x}$\;
    	}
    }
\end{algorithm}
\vspace{-0.5cm}

The rationale behind the RI mechanism is then to restart the algorithm from a partially random solution, i.e. a solution that is randomly generated but still inherits part of the variables from the current best. This way the restart is not entirely disruptive, but preserves at least a block of (an expected number of) $D\alpha$ variables. This partial inheritance allows the algorithm to keep some information from one restart and the next one, but also to escape from local optima. From this point of view, the RI mechanism shares some resemblance with the Iterated Local Search (ILS) methods \cite{ILS2010}, that try to apply a small perturbation to the best-so-far solution during restart (in fact, as small as possible to not disrupt it much, but as large as needed to allow the local search to converge to a different local optimum). However, ILS has been especially designed for (and applied to) combinatorial optimization, while here we focus on continuous optimization.

\section{Numerical Results}\label{results}
In the following we present the numerical results obtained on the CEC 2014 benchmark \cite{cec2014}. This benchmark is composed of $30$ functions, with different properties in terms of separability, ill-conditioning, and landscape structure. In particular, it is worth noting that except for $f_8$ (Shifted Rastrigin's Function) and $f_{10}$ (Shifted Schwefel's Function), all CEC 2014 benchmark functions are non-separable. Therefore this benchmark is particularly suited for testing the performance of optimization algorithms on non-separable problems.

We considered the following four real-valued compact algorithms, with the parametrization proposed in their original papers:
\begin{itemize}
\item cDE ``light'' \cite{cDELight} with exponential crossover and parameters: $N_p= 300$, $F = 0.5$, and $\alpha_m = 0.25$;
\item rcGA \cite{rcGA}, with persistent elitism and parameters: $N_p = 300$;
\item cPSO \cite{cPSO}, with parameters: $N_p = 300$, $\phi_1 = 0.2$, $\phi_2 = 0.07$, $\phi_3 = 3.74$, $\gamma_1 = 1$, and $\gamma_2 = 1$;
\item cBFO \cite{cBFO}, with parameters: $N_p = 300$, $C_i = 0.1$, and $N_s = 4$.
\end{itemize}

As for the corresponding versions with RI (dubbed, respectively as RIcDE, RIrcGA, RIcPSO and RIcBFO), the same parametrization was kept for the compact optimization process while the RI component was parametrized with $\alpha=0.05$ (such that only $5\%$ of the variables are inherited, on average, from the current best). A number of fitness function calls equal to $25\%$ of the total computational budget was assigned to execute the compact algorithm after each restart. It should be noted that these are the only two parameters of the RI mechanism and they were empirically set after having observed their effect in preliminary experiments.

Furthermore, in order to assess the effect of the RI mechanism w.r.t. a simple random restart without any form of inheritance, we included in our experimental setup also four variants of the same compact algorithms where the restart was applied, with the same period of the RI variants ($25\%$ of the total computational budget), by simply applying a uniform re-sampling of a new solution $\bf{x}$ within the search space, and restarting the compact algorithm by setting $\boldsymbol{\mu}=\bf{x}$. We dub these compact algorithms with random restart, respectively, as RecDE, RercGA, RecPSO and RecBFO.

Finally, to provide a baseline for all the compact algorithms with/without restart tested in this paper, we evaluated the performance of a simple Random Walk (RW) algorithm where at each step a new solution is generated by applying a uniform re-sampling within the search space. From our numerical results (see Table \ref{holm-test-all}) it can be seen that its performance is -as expected- considerably worse than any of the compact algorithms considered in the experiments, thus highlighting that the ``compact'' logic is more than a mere random sampling and performs significantly better w.r.t. pure uniform random searches.

To assess the scalability of all the algorithms, we performed experiments in 10, 50 and 100 dimensions. Thus, the total experimental setup consists of 13 algorithms (4 compact algorithms, 4 RI variants, 4 variants with random restart, and RW) and $30 \times 3 = 90$ optimization problems (i.e. 30 functions each tested in three different dimensionalities). On each benchmark function, each algorithm was executed for $30$ independent runs, to collect statistics on the fitness values obtained in each run at the end of the allotted computational budget. Each run was executed for a total budget of $5000 \times D$ function evaluations, being $D$ the problem dimension.

In the following, for the sake of brevity we will show only a compact representation of the main experimental results (detailed numerical results are available in the Appendix). For that, we will use the sequentially rejective Holm-Bonferroni procedure \cite{bib:Garcia2008b,bib:Holm1979}. This procedure consists of the following: considering $N_{TP}$ test problems (in our case, $90$) and $N_A$ optimization algorithms, the performance obtained by each algorithm on each problem is computed. This is measured as average of the best fitness values obtained by the algorithm on that problem over multiple (in our case, $30$) independent runs, at the end of the computational budget (in our case, $5000 \times D$ function evaluations). Then, for each problem a score $R_i$ is assigned to each algorithm, being $N_A$ the score of the algorithm displaying the best performance (i.e., assuming minimization, the minimum average of the fitness values) on that problem, $N_A-1$ the score of the second best, and so on. The algorithm displaying the worst performance scores $1$. These scores are then averaged, for each algorithm, over the whole set of $N_{TP}$ test problems. The algorithms are sorted on the basis of these average scores. Indicating with $R_0$ the rank of an algorithm taken as reference, and with $R_j$ for $j = 1,\dots,N_A-1$ the rank of the remaining algorithms, the values $z_j$ are calculated as:
\begin{equation}
z_j = \frac{R_j - R_0}{\sqrt{\frac{N_A(N_A+1)}{6N_{TP}}}}.
\end{equation}
By means of the $z_j$ values, the corresponding cumulative normal distribution values $p_j$ are derived. These $p_j$ values are then compared to the corresponding $\delta /j$ where $\delta$ is the confidence interval, set to $0.05$: if $p_j > \delta /j$, the null-hypothesis (that the algorithm taken as reference has the same performance as the j-th algorithm) is accepted, otherwise is rejected as well as all the subsequent tests.

Let us first consider, for each compact algorithm, how the corresponding algorithms with RI and random restart perform w.r.t. the original compact algorithm without restart. Tables \ref{holm-test-cDE}-\ref{holm-test-cBFO} show, respectively, the results of the Holm-Bonferroni procedure (in this case with $N_A=3$) on cDE, rcGA, cPSO and cBFO based algorithms. The tables display the ranks, $z_j$ values, $p_j$ values, and corresponding $\delta/j$ obtained by this procedure. In each case we considered as reference algorithm the corresponding algorithm with RI, whose rank is shown in parenthesis in each table caption. Moreover, we indicate in each table whether the null-hypothesis (that the algorithm taken as reference has the same performance as each other algorithm in the corresponding table row) is accepted or not.
\vspace{-0.5cm}
\begin{table}[ht!]
\centering
\caption{Holm-Bonferroni procedure (reference: RIcDE, Rank = 2.63e+00)}\label{holm-test-cDE}
\begin{tabular}{c|c|c|c|c|c|c}
\hline\hline
$j$ & Optimizer & Rank & $z_j$ & $p_j$ & $\delta/j$ & Hypothesis\\
\hline
1 & RecDE & 2.33e+00 & -2.85e+00 & 2.21e-03 & 5.00e-02 & Rejected\\
2 & cDE & 1.03e+00 & -1.52e+01 & 2.44e-52 & 2.50e-02 & Rejected\\
\hline\hline
\end{tabular}
\end{table}

\vspace{-1cm}
\begin{table}[ht!]
\centering
\caption{Holm-Bonferroni procedure (reference: RIrcGA, Rank = 2.53e+00)}\label{holm-test-rcGA}
\begin{tabular}{c|c|c|c|c|c|c}
\hline\hline
$j$ & Optimizer & Rank & $z_j$ & $p_j$ & $\delta/j$ & Hypothesis\\
\hline
1 & RercGA & 2.47e+00 & -6.32e-01 & 2.64e-01 & 5.00e-02 & Accepted\\
2 & rcGA & 1.00e+00 & -1.45e+01 & 3.07e-48 & 2.50e-02 & Rejected\\
\hline\hline
\end{tabular}
\end{table}

\vspace{-1cm}
\begin{table}[ht!]
\centering
\caption{Holm-Bonferroni procedure (reference: RIcPSO, Rank = 1.99e+00)}\label{holm-test-cPSO}
\begin{tabular}{c|c|c|c|c|c|c}
\hline\hline
$j$ & Optimizer & Rank & $z_j$ & $p_j$ & $\delta/j$ & Hypothesis\\
\hline
1 & RecPSO & 2.01e+00 & 2.11e-01 & 5.83e-01 & 5.00e-02 & Accepted\\
2 & cPSO & 2.00e+00 & 1.05e-01 & 5.42e-01 & 2.50e-02 & Accepted\\
\hline\hline
\end{tabular}
\end{table}

\vspace{-1cm}
\begin{table}[ht!]
\centering
\caption{Holm-Bonferroni procedure (reference: RIcBFO, Rank = 1.84e+00)}\label{holm-test-cBFO}
\begin{tabular}{c|c|c|c|c|c|c}
\hline\hline
$j$ & Optimizer & Rank & $z_j$ & $p_j$ & $\delta/j$ & Hypothesis\\
\hline
1 & RecBFO & 1.88e+00 & 3.16e-01 & 6.24e-01 & 5.00e-02 & Accepted\\
2 & cBFO & 1.54e+00 & -2.85e+00 & 2.21e-03 & 2.50e-02 & Rejected\\
\hline\hline
\end{tabular}
\end{table}

\vspace{-0.5cm}

From these Holm-Bonferroni procedures, we can observe that, except for the case of cPSO (where, quite surprisingly, cPSO shows the same performance as the corresponding algorithms with RI and random restart) in all other cases the algorithms with RI score a better rank than their corresponding compact algorithms. It is also interesting to note that, while in the case of cDE RIcDE performs better also than RecDE, on the other compact algorithms it results that the RI variant is statistically equivalent to the variant with random restart (note that the null-hypothesis is accepted in those cases). This equivalence between RI and random restart in the case of rcGA, cPSO and cBFO might be due to parametrization used for RI (number of restarts and number of variables inherited from the current best), as well as the different algorithmic logics used by these algorithms compared to cDE. In general though, these observations demonstrate that the use of restarts, and, especially in the case of cDE, the use of RI is beneficial in terms of optimization performance.

Finally, we provide an overall comparison of all the $12$ compact optimization algorithms, in addition to the Random Walk algorithm. The resulting Holm-Bonferroni procedure is reported in Table \ref{holm-test-all}, where RIcDE is considered as reference algorithm (as it shows the highest rank) and $N_A=13$. In this case, except for RecDE, all the hypotheses are sequentially rejected, meaning that when all the algorithms are considered together, RIcDE is statistically equivalent to RecDE (although it shows a slightly higher rank), but it shows a statistically better performance (on average, on the entire set of tested problems) than all other algorithms under study. As expected, the Random Walk algorithm performs worse than all other papers. Moreover, the rank shows that each compact algorithm with RI (or random restart) performs better (on average) than the corresponding compact algorithm algorithm, confirming the fact that the RI component is beneficial to all the compact algorithms considered in our experimentation.
\vspace{-0.5cm}
\begin{table}[ht!]
\centering
\caption{Holm-Bonferroni procedure (reference: RIcDE, Rank = 1.09e+01)}\label{holm-test-all}
\begin{tabular}{c|c|c|c|c|c|c}
\hline\hline
$j$ & Optimizer & Rank & $z_j$ & $p_j$ & $\delta/j$ & Hypothesis\\
\hline
1 & RecDE & 1.06e+01 & -6.41e-01 & 2.61e-01 & 5.00e-02 & Accepted\\
2 & RecBFO & 8.56e+00 & -4.38e+00 & 5.86e-06 & 2.50e-02 & Rejected\\
3 & RIcBFO & 8.52e+00 & -4.44e+00 & 4.40e-06 & 1.67e-02 & Rejected\\
4 & RIrcGA & 8.38e+00 & -4.71e+00 & 1.22e-06 & 1.25e-02 & Rejected\\
5 & RercGA & 8.30e+00 & -4.86e+00 & 5.93e-07 & 1.00e-02 & Rejected\\
6 & cBFO & 8.07e+00 & -5.29e+00 & 6.04e-08 & 8.33e-03 & Rejected\\
7 & cDE & 8.03e+00 & -5.35e+00 & 4.30e-08 & 7.14e-03 & Rejected\\
8 & rcGA & 5.68e+00 & -9.74e+00 & 1.05e-22 & 6.25e-03 & Rejected\\
9 & RecPSO & 3.97e+00 & -1.29e+01 & 1.73e-38 & 5.56e-03 & Rejected\\
10 & RIcPSO & 3.97e+00 & -1.29e+01 & 1.73e-38 & 5.56e-03 & Rejected\\
11 & cPSO & 3.89e+00 & -1.31e+01 & 2.61e-39 & 5.00e-03 & Rejected\\
12 & RW & 1.43e+00 & -1.76e+01 & 6.80e-70 & 4.55e-03 & Rejected\\
\hline\hline
\end{tabular}
\end{table}

\vspace{-1cm}

\section{Conclusions}\label{conclusions}
In this paper we have presented an algorithmic scheme for solving continuous optimization problems on devices characterized by limited memory. The proposed scheme is based on a combination of a compact algorithm with a restart mechanism based on Re-Sampled Inheritance (RI). We tested this scheme on four different compact algorithms presented in the literature (namely: cDE, rcGA, cPSO, and cBFO) and performed numerical experiments on a broad range of benchmark functions in several dimensionalities. Our experiments show that the use of RI consistently enhances the performances of compact algorithms, still keeping a limited usage of memory. In addition to that, we noted that among the tested algorithms the best performance was obtained by cDE with Re-Sampled Inheritance.

In future works, we will further investigate the effect of the parametrization on the proposed compact algorithms with Re-Sampled Inheritance, focusing in particular on the influence of the number of restarts, as well as the number of variables inherited from the best individual at each restart. We will also investigate alternative inheritance mechanisms, for instance based on binomial crossover or exponential crossover on shuffled variables.


\vspace{0.2cm}
\noindent
\begin{tabular}{p{0.15\linewidth} p{0.85\linewidth}}
\raisebox{-0.8cm}{\includegraphics[height=.95cm]{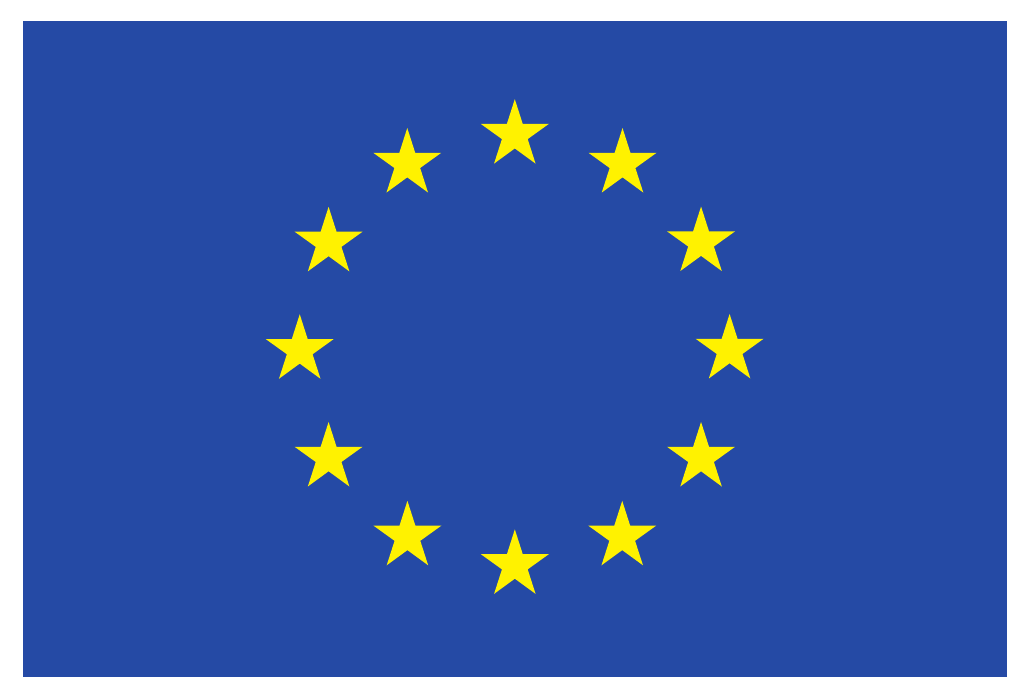}}
&
{\textbf{Acknowledgments.} \small\fontfamily{arial}\selectfont This project has received funding from the European Union's Horizon 2020 research and innovation programme under grant agreement No 665347.}
\end{tabular}
\vspace{-0.7cm}

%
%
\bibliographystyle{splncs03}
\bibliography{iacca}

\begin{thebibliography}{10}
\providecommand{\url}[1]{\texttt{#1}}
\providecommand{\urlprefix}{URL }

\bibitem{compactOptimization}
Neri, F., Iacca, G., Mininno, E.: Compact Optimization, pp. 337--364. Springer
  (2013)

\bibitem{EDAbook}
Larra\~{n}aga, P., Lozano, J.A.: Estimation of Distribution Algorithms: A New
  Tool for Evolutionary Computation. Kluwer Academic Publishers (2001)

\bibitem{cGA}
Harik, G.R., Lobo, F.G., Goldberg, D.E.: The compact genetic algorithm. IEEE
  transactions on evolutionary computation  3(4),  287--297 (1999)

\bibitem{SG}
Corno, F., Reorda, M.S., Squillero, G.: The selfish gene algorithm: A new
  evolutionary optimization strategy. In: ACM Symposium on Applied Computing.
  pp. 349--355 (1998)

\bibitem{cGAElitism}
Ahn, C.W., Ramakrishna, R.S.: Elitism-based compact genetic algorithms. IEEE
  Transactions on Evolutionary Computation  7(4),  367--385 (2003)

\bibitem{familycGA}
Gallagher, J.C., Vigraham, S., Kramer, G.: A family of compact genetic
  algorithms for intrinsic evolvable hardware. IEEE Transactions on
  evolutionary computation  8(2),  111--126 (2004)

\bibitem{rcGA}
Mininno, E., Cupertino, F., Naso, D.: Real-valued compact genetic algorithms
  for embedded microcontroller optimization. IEEE Transactions on Evolutionary
  Computation  12(2),  203--219 (2008)

\bibitem{cDE}
Mininno, E., Neri, F., Cupertino, F., Naso, D.: Compact differential evolution.
  IEEE Transactions on Evolutionary Computation  15(1),  32--54 (2011)

\bibitem{globalcDE}
Iacca, G., Mallipeddi, R., Mininno, E., Neri, F., Suganthan, P.N.: Global
  supervision for compact differential evolution. In: IEEE Symposium on
  Differential Evolution. pp. 1--8 (2011)

\bibitem{superfitcDE}
Iacca, G., Mallipeddi, R., Mininno, E., Neri, F., Suganthan, P.N.: Super-fit
  and population size reduction in compact differential evolution. In: IEEE
  Workshop on Memetic Computing. pp. 1--8 (2011)

\bibitem{cDELight}
Iacca, G., Caraffini, F., Neri, F.: Compact differential evolution light: High
  performance despite limited memory requirement and modest computational
  overhead. Journal of Computer Science and Technology  27(5),  1056--1076
  (2012)

\bibitem{coDE}
Iacca, G., Mininno, E., Neri, F.: Composed compact differential evolution.
  Evolutionary Intelligence  4(1),  17--29 (2011)

\bibitem{obcDE}
Iacca, G., Neri, F., Mininno, E.: Opposition-based learning in compact
  differential evolution. In: Applications of Evolutionary Computation. pp.
  264--273 (2011)

\bibitem{nacDE}
Iacca, G., Neri, F., Mininno, E.: Noise analysis compact differential
  evolution. International Journal of Systems Science  43(7),  1248--1267
  (2012)

\bibitem{CMcDE}
Jewajinda, Y.: Covariance matrix compact differential evolution for embedded
  intelligence. In: IEEE Region 10 Symposium. pp. 349--354 (2016)

\bibitem{ensemblecDE}
Mallipeddi, R., Iacca, G., Suganthan, P.N., Neri, F., Mininno, E.: Ensemble
  strategies in compact differential evolution. In: IEEE Congress on
  Evolutionary Computation. pp. 1972--1977 (2011)

\bibitem{McDE}
Neri, F.: Memetic compact differential evolution for cartesian robot control.
  IEEE Computational Intelligence Magazine  5(2),  54--65 (2010)

\bibitem{disturbedcDE}
Neri, F., Iacca, G., Mininno, E.: Disturbed exploitation compact differential
  evolution for limited memory optimization problems. Information Sciences
  181(12),  2469 -- 2487 (2011)

\bibitem{cPSO}
Neri, F., Mininno, E., Iacca, G.: Compact particle swarm optimization.
  Information Sciences  239,  96 -- 121 (2013)

\bibitem{cBFO}
Iacca, G., Neri, F., Mininno, E.: Compact bacterial foraging optimization. In:
  Swarm and Evolutionary Computation. pp. 84--92 (2012)

\bibitem{cTLBO}
Yang, Z., Li, K., Guo, Y.: A new compact teaching-learning-based optimization
  method. In: International Conference on Intelligent Computing. pp. 717--726
  (2014)

\bibitem{rcTLBO}
Yang, Z., Li, K., Guo, Y., Ma, H., Zheng, M.: Compact real-valued
  teaching-learning based optimization with the applications to neural network
  training. Knowledge-Based Systems  (2018)

\bibitem{EcABC}
Banitalebi, A., Aziz, M.I.A., Bahar, A., Aziz, Z.A.: Enhanced compact
  artificial bee colony. Information Sciences  298,  491--511 (2015)

\bibitem{cABC}
Dao, T.K., Chu, S.C., Shieh, C.S., Horng, M.F.: Compact artificial bee colony.
  In: International Conference on Industrial, Engineering and Other
  Applications of Applied Intelligent Systems. pp. 96--105 (2014)

\bibitem{cFPA}
Dao, T.K., Pan, T.S., Nguyen, T.T., Chu, S.C., Pan, J.S.: A compact flower
  pollination algorithm optimization. In: International Conference on Computing
  Measurement Control and Sensor Network. pp. 76--79 (2016)

\bibitem{cDELightRobot}
Iacca, G., Caraffini, F., Neri, F., Mininno, E.: Robot base disturbance
  optimization with compact differential evolution light. In: Applications of
  Evolutionary Computation. pp. 285--294 (2012)

\bibitem{cABC-WSN}
Dao, T.K., Pan, T.S., Nguyen, T.T., Chu, S.C.: A compact artificial bee colony
  optimization for topology control scheme in wireless sensor networks. Journal
  of Information Hiding and Multimedia Signal Processing  6(2),  297--310
  (2015)

\bibitem{RICaraffini2013a}
Caraffini, F., Iacca, G., Neri, F., Picinali, L., Mininno, E.: {A CMA-ES
  super-fit scheme for the re-sampled inheritance search}. In: IEEE Congress on
  Evolutionary Computation. pp. 1123--1130 (2013)

\bibitem{RICaraffini2013b}
Caraffini, F., Neri, F., Passow, B.N., Iacca, G.: Re-sampled inheritance
  search: high performance despite the simplicity. Soft Computing  17(12),
  2235--2256 (2013)

\bibitem{LEGO2018RICMAES}
Caraffini, F., Iacca, G., Yaman, A.: {Improving (1+1) Covariance Matrix
  Adaptation Evolution Strategy: a simple yet efficient approach}. In:
  International Global Optimization Workshop (to appear 2018)

\bibitem{cec2014}
Liang, J., Qu, B., Suganthan, P.: {Problem definitions and evaluation criteria
  for the CEC 2014 special session and competition on single objective
  real-parameter numerical optimization}. Computational Intelligence
  Laboratory, Zhengzhou University, Zhengzhou China and Technical Report,
  Nanyang Technological University, Singapore  (2013)

\bibitem{ILS2010}
Louren{\c{c}}o, H.R., Martin, O.C., St{\"u}tzle, T.: Iterated local search:
  Framework and applications. In: Handbook of metaheuristics, pp. 363--397.
  Springer (2010)

\bibitem{bib:Garcia2008b}
Garcia, S., Fernandez, A., Luengo, J., Herrera, F.: {A study of statistical
  techniques and performance measures for genetics-based machine learning:
  accuracy and interpretability}. Soft Computing  13(10),  959--977 (2008)

\bibitem{bib:Holm1979}
Holm, S.: {A simple sequentially rejective multiple test procedure}.
  Scandinavian Journal of Statistics  6(2),  65--70 (1979)

\bibitem{bib:Wilcoxon1945}
Wilcoxon, F.: {Individual comparisons by ranking methods}. Biometrics Bulletin
  1(6),  80--83 (1945)

\end{thebibliography}

\newpage
\section*{Appendix: Extended results}\label{extended}
Tables \ref{tab:cDE10D}-\ref{tab:cBFO100D} report the detailed numerical results on each problem, in terms of: 1) average error w.r.t. the known optimal fitness $\pm$ std. dev.; 2) pairwise comparisons -according to the non-parametric Wilcoxon Rank-Sum test \cite{bib:Wilcoxon1945}- between one algorithm (in each table, the RI variant), taken as reference, and the other algorithms shown in the table. More specifically:
\begin{itemize}
    \item Tables \ref{tab:cDE10D}, \ref{tab:cDE50D} and \ref{tab:cDE100D} report the results in 10, 50 and 100 dimensions, respectively, of the cDE-based algorithms, i.e. RIcDE (taken as reference), RecDE, and cDE. 
    \item Tables \ref{tab:rcGA10D}, \ref{tab:rcGA50D} and \ref{tab:rcGA100D} report the results in 10, 50 and 100 dimensions, respectively, of the rcGA-based algorithms, i.e. RIrcGA (taken as reference), RercGA, and rcGA.
    \item Tables \ref{tab:cPSO10D}, \ref{tab:cPSO50D} and \ref{tab:cPSO100D} report the results in 10, 50 and 100 dimensions, respectively, of the cPSO-based algorithms, i.e. RIcPSO (taken as reference), RecPSO, and cPSO.
    \item Tables \ref{tab:cBFO10D}, \ref{tab:cBFO50D} and \ref{tab:cBFO100D} report the results in 10, 50 and 100 dimensions, respectively, of the cBFO-based algorithms, i.e. RIcBFO (taken as reference), RecBFO, and cBFO.
\end{itemize}
In each table, the boldface indicates the algorithm that obtains the minimum average error on each tested benchmark function. The result of each pairwise Wilcoxon Rank-Sum test is indicated as `+', `-', or `=', representing, respectively, the fact that the reference algorithm for each table is better than/worse than/equal to the algorithm corresponding to the focal column label. In all the pairwise tests, we considered a confidence interval of $0.05$.

Finally, Table \ref{tab:RW} shows the baseline results obtained by Random Walk on all the benchmark problems in 10, 50 and 100 dimensions,

\begin{table}[p]
\caption{Average error $\pm$ standard deviation and Wilcoxon Rank-Sum Test (reference: RIcDE) for RIcDE against RecDE and cDE on CEC2014~\cite{cec2014} in $10$D.}\label{tab:cDE10D}
\centering
\begin{tabular}{l|r@{$\,\pm\,$}l|c|r@{$\,\pm\,$}l|c|r@{$\,\pm\,$}l|c}
\hline\hline
              &      \multicolumn{3}{c|}{RIcDE}    &   \multicolumn{3}{c|}{RecDE}&   \multicolumn{3}{c}{cDE}\\
\hline
$f_{1}$ & \multicolumn{3}{c|}{$1.38e+05\,\pm\,1.55e+05$} & $2.08e+05$ & $2.70e+05$ & = & $\mathbf{6.33e+04}$ & $\mathbf{5.38e+04}$ & -\\
$f_{2}$ & \multicolumn{3}{c|}{$\mathbf{8.96e+02}\,\pm\,\mathbf{1.22e+03}$} & $1.22e+03$ & $1.67e+03$ & = & $4.41e+03$ & $3.13e+03$ & +\\
$f_{3}$ & \multicolumn{3}{c|}{$\mathbf{6.92e+02}\,\pm\,\mathbf{6.26e+02}$} & $1.40e+03$ & $1.80e+03$ & = & $7.30e+03$ & $6.04e+03$ & +\\
$f_{4}$ & \multicolumn{3}{c|}{$\mathbf{2.76e+00}\,\pm\,\mathbf{8.60e+00}$} & $4.57e+00$ & $1.02e+01$ & = & $1.34e+01$ & $1.63e+01$ & =\\
$f_{5}$ & \multicolumn{3}{c|}{$\mathbf{2.00e+01}\,\pm\,\mathbf{3.23e-03}$} & $2.00e+01$ & $1.49e-03$ & = & $2.00e+01$ & $4.12e-05$ & -\\
$f_{6}$ & \multicolumn{3}{c|}{$\mathbf{1.16e+00}\,\pm\,\mathbf{7.62e-01}$} & $1.34e+00$ & $6.83e-01$ & = & $2.60e+00$ & $1.46e+00$ & +\\
$f_{7}$ & \multicolumn{3}{c|}{$9.91e-02\,\pm\,4.62e-02$} & $\mathbf{9.09e-02}$ & $\mathbf{3.97e-02}$ & = & $1.72e-01$ & $1.10e-01$ & +\\
\hline\hline
\end{tabular}
\end{table}

\begin{table}[p]
\caption{Average error $\pm$ standard deviation and Wilcoxon Rank-Sum Test (reference: RIcDE) for RIcDE against RecDE and cDE on CEC2014~\cite{cec2014} in $50$D.}\label{tab:cDE50D}
\centering
\begin{tabular}{l|r@{$\,\pm\,$}l|c|r@{$\,\pm\,$}l|c|r@{$\,\pm\,$}l|c}
\hline\hline
              &      \multicolumn{3}{c|}{RIcDE}    &   \multicolumn{3}{c|}{RecDE}&   \multicolumn{3}{c}{cDE}\\
\hline
$f_{1}$ & \multicolumn{3}{c|}{$\mathbf{7.12e+06}\,\pm\,\mathbf{2.83e+06}$} & $9.53e+06$ & $3.34e+06$ & + & $1.20e+07$ & $4.83e+06$ & +\\
$f_{2}$ & \multicolumn{3}{c|}{$\mathbf{1.67e+03}\,\pm\,\mathbf{3.41e+03}$} & $2.21e+03$ & $3.11e+03$ & = & $6.41e+03$ & $8.41e+03$ & +\\
$f_{3}$ & \multicolumn{3}{c|}{$\mathbf{1.86e+04}\,\pm\,\mathbf{7.37e+03}$} & $1.91e+04$ & $7.41e+03$ & = & $2.91e+04$ & $1.00e+04$ & +\\
$f_{4}$ & \multicolumn{3}{c|}{$\mathbf{1.24e+02}\,\pm\,\mathbf{2.19e+01}$} & $1.32e+02$ & $3.13e+01$ & = & $1.88e+02$ & $5.29e+01$ & +\\
$f_{5}$ & \multicolumn{3}{c|}{$2.00e+01\,\pm\,9.92e-06$} & $\mathbf{2.00e+01}$ & $\mathbf{3.07e-06}$ & = & $2.00e+01$ & $1.08e-02$ & +\\
$f_{6}$ & \multicolumn{3}{c|}{$3.25e+01\,\pm\,2.81e+00$} & $\mathbf{3.23e+01}$ & $\mathbf{3.42e+00}$ & = & $3.72e+01$ & $4.20e+00$ & +\\
$f_{7}$ & \multicolumn{3}{c|}{$\mathbf{2.35e-02}\,\pm\,\mathbf{1.61e-02}$} & $3.57e-02$ & $3.13e-02$ & = & $6.31e-02$ & $6.63e-02$ & +\\
\hline\hline
\end{tabular}
\end{table}

\begin{table}[p]
\caption{Average error $\pm$ standard deviation and Wilcoxon Rank-Sum Test (reference: RIcDE) for RIcDE against RecDE and cDE on CEC2014~\cite{cec2014} in $100$D.}\label{tab:cDE100D}
\centering
\begin{tabular}{l|r@{$\,\pm\,$}l|c|r@{$\,\pm\,$}l|c|r@{$\,\pm\,$}l|c}
\hline\hline
              &      \multicolumn{3}{c|}{RIcDE}    &   \multicolumn{3}{c|}{RecDE}&   \multicolumn{3}{c}{cDE}\\
\hline
$f_{1}$ & \multicolumn{3}{c|}{$\mathbf{8.59e+07}\,\pm\,\mathbf{2.56e+07}$} & $1.20e+08$ & $3.37e+07$ & + & $1.72e+08$ & $5.11e+07$ & +\\
$f_{2}$ & \multicolumn{3}{c|}{$\mathbf{4.40e+07}\,\pm\,\mathbf{3.21e+07}$} & $6.24e+08$ & $3.14e+08$ & + & $7.45e+08$ & $4.18e+08$ & +\\
$f_{3}$ & \multicolumn{3}{c|}{$\mathbf{3.68e+04}\,\pm\,\mathbf{1.36e+04}$} & $4.89e+04$ & $1.57e+04$ & + & $5.74e+04$ & $1.55e+04$ & +\\
$f_{4}$ & \multicolumn{3}{c|}{$\mathbf{4.44e+02}\,\pm\,\mathbf{6.92e+01}$} & $6.54e+02$ & $1.12e+02$ & + & $7.01e+02$ & $1.21e+02$ & +\\
$f_{5}$ & \multicolumn{3}{c|}{$\mathbf{2.00e+01}\,\pm\,\mathbf{2.54e-02}$} & $2.01e+01$ & $2.41e-02$ & + & $2.01e+01$ & $5.13e-02$ & +\\
$f_{6}$ & \multicolumn{3}{c|}{$9.57e+01\,\pm\,5.43e+00$} & $\mathbf{9.54e+01}$ & $\mathbf{5.12e+00}$ & = & $1.02e+02$ & $7.54e+00$ & +\\
$f_{7}$ & \multicolumn{3}{c|}{$\mathbf{1.34e+00}\,\pm\,\mathbf{5.34e-01}$} & $6.82e+00$ & $2.56e+00$ & + & $8.77e+00$ & $6.07e+00$ & +\\
\hline\hline
\end{tabular}
\end{table}


\begin{table}[p]
\caption{Average error $\pm$ standard deviation and Wilcoxon Rank-Sum Test (reference: RIrcGA) for RIrcGA against RercGA and rcGA on CEC2014~\cite{cec2014} in $10$D.}\label{tab:rcGA10D}
\centering
\begin{tabular}{l|r@{$\,\pm\,$}l|c|r@{$\,\pm\,$}l|c|r@{$\,\pm\,$}l|c}
\hline\hline
              &      \multicolumn{3}{c|}{RIrcGA}    &   \multicolumn{3}{c|}{RercGA}&   \multicolumn{3}{c}{rcGA}\\
\hline
$f_{1}$ & \multicolumn{3}{c|}{$\mathbf{1.38e+05}\,\pm\,\mathbf{7.91e+04}$} & $1.75e+05$ & $1.87e+05$ & = & $4.50e+05$ & $1.74e+06$ & =\\
$f_{2}$ & \multicolumn{3}{c|}{$\mathbf{1.06e+03}\,\pm\,\mathbf{1.58e+03}$} & $1.22e+03$ & $1.82e+03$ & = & $4.37e+03$ & $3.50e+03$ & +\\
$f_{3}$ & \multicolumn{3}{c|}{$3.71e+03\,\pm\,2.81e+03$} & $\mathbf{3.22e+03}$ & $\mathbf{2.89e+03}$ & = & $8.60e+03$ & $6.62e+03$ & +\\
$f_{4}$ & \multicolumn{3}{c|}{$1.67e+01\,\pm\,1.55e+01$} & $\mathbf{1.54e+01}$ & $\mathbf{1.55e+01}$ & = & $2.70e+01$ & $1.42e+01$ & +\\
$f_{5}$ & \multicolumn{3}{c|}{$2.00e+01\,\pm\,8.08e-04$} & $\mathbf{2.00e+01}$ & $\mathbf{3.98e-03}$ & = & $2.00e+01$ & $8.30e-05$ & +\\
$f_{6}$ & \multicolumn{3}{c|}{$\mathbf{1.14e+00}\,\pm\,\mathbf{7.46e-01}$} & $1.30e+00$ & $7.76e-01$ & = & $2.87e+00$ & $1.76e+00$ & +\\
$f_{7}$ & \multicolumn{3}{c|}{$8.69e-02\,\pm\,3.11e-02$} & $\mathbf{7.90e-02}$ & $\mathbf{4.00e-02}$ & = & $1.77e-01$ & $1.19e-01$ & +\\
\hline\hline
\end{tabular}
\end{table}

\begin{table}[p]
\caption{Average error $\pm$ standard deviation and Wilcoxon Rank-Sum Test (reference: RIrcGA) for RIrcGA against RercGA and rcGA on CEC2014~\cite{cec2014} in $50$D.}\label{tab:rcGA50D}
\centering
\begin{tabular}{l|r@{$\,\pm\,$}l|c|r@{$\,\pm\,$}l|c|r@{$\,\pm\,$}l|c}
\hline\hline
              &      \multicolumn{3}{c|}{RIrcGA}    &   \multicolumn{3}{c|}{RercGA}&   \multicolumn{3}{c}{rcGA}\\
\hline
$f_{1}$ & \multicolumn{3}{c|}{$\mathbf{5.60e+07}\,\pm\,\mathbf{2.15e+07}$} & $5.72e+07$ & $1.84e+07$ & = & $1.02e+08$ & $5.26e+07$ & +\\
$f_{2}$ & \multicolumn{3}{c|}{$\mathbf{1.16e+09}\,\pm\,\mathbf{4.77e+08}$} & $1.19e+09$ & $5.38e+08$ & = & $2.57e+09$ & $1.55e+09$ & +\\
$f_{3}$ & \multicolumn{3}{c|}{$\mathbf{1.30e+05}\,\pm\,\mathbf{2.86e+04}$} & $1.44e+05$ & $2.34e+04$ & + & $1.80e+05$ & $3.61e+04$ & +\\
$f_{4}$ & \multicolumn{3}{c|}{$4.02e+02\,\pm\,9.70e+01$} & $\mathbf{4.02e+02}$ & $\mathbf{9.18e+01}$ & = & $5.82e+02$ & $2.98e+02$ & +\\
$f_{5}$ & \multicolumn{3}{c|}{$\mathbf{2.02e+01}\,\pm\,\mathbf{7.59e-02}$} & $2.02e+01$ & $8.73e-02$ & = & $2.03e+01$ & $1.16e-01$ & +\\
$f_{6}$ & \multicolumn{3}{c|}{$3.99e+01\,\pm\,3.98e+00$} & $\mathbf{3.95e+01}$ & $\mathbf{2.82e+00}$ & = & $4.61e+01$ & $5.47e+00$ & +\\
$f_{7}$ & \multicolumn{3}{c|}{$\mathbf{1.45e+01}\,\pm\,\mathbf{8.61e+00}$} & $1.57e+01$ & $5.93e+00$ & = & $3.09e+01$ & $2.04e+01$ & +\\
\hline\hline
\end{tabular}
\end{table}

\begin{table}[p]
\caption{Average error $\pm$ standard deviation and Wilcoxon Rank-Sum Test (reference: RIrcGA) for RIrcGA against RercGA and rcGA on CEC2014~\cite{cec2014} in $100$D.}\label{tab:rcGA100D}
\centering
\begin{tabular}{l|r@{$\,\pm\,$}l|c|r@{$\,\pm\,$}l|c|r@{$\,\pm\,$}l|c}
\hline\hline
              &      \multicolumn{3}{c|}{RIrcGA}    &   \multicolumn{3}{c|}{RercGA}&   \multicolumn{3}{c}{rcGA}\\
\hline
$f_{1}$ & \multicolumn{3}{c|}{$\mathbf{5.11e+08}\,\pm\,\mathbf{1.31e+08}$} & $5.59e+08$ & $1.16e+08$ & = & $8.10e+08$ & $1.88e+08$ & +\\
$f_{2}$ & \multicolumn{3}{c|}{$4.57e+10\,\pm\,1.15e+10$} & $\mathbf{4.39e+10}$ & $\mathbf{1.06e+10}$ & = & $5.35e+10$ & $1.99e+10$ & =\\
$f_{3}$ & \multicolumn{3}{c|}{$\mathbf{4.19e+05}\,\pm\,\mathbf{4.22e+04}$} & $4.19e+05$ & $5.62e+04$ & = & $4.83e+05$ & $8.37e+04$ & +\\
$f_{4}$ & \multicolumn{3}{c|}{$4.95e+03\,\pm\,1.05e+03$} & $\mathbf{4.84e+03}$ & $\mathbf{1.21e+03}$ & = & $6.43e+03$ & $2.50e+03$ & =\\
$f_{5}$ & \multicolumn{3}{c|}{$2.06e+01\,\pm\,8.27e-02$} & $\mathbf{2.06e+01}$ & $\mathbf{7.70e-02}$ & = & $2.07e+01$ & $1.05e-01$ & +\\
$f_{6}$ & \multicolumn{3}{c|}{$1.07e+02\,\pm\,6.45e+00$} & $\mathbf{1.05e+02}$ & $\mathbf{5.89e+00}$ & = & $1.14e+02$ & $8.21e+00$ & +\\
$f_{7}$ & \multicolumn{3}{c|}{$\mathbf{3.62e+02}\,\pm\,\mathbf{7.12e+01}$} & $3.72e+02$ & $6.64e+01$ & = & $5.34e+02$ & $1.31e+02$ & +\\
\hline\hline
\end{tabular}
\end{table}


\begin{table}[p]
\caption{Average error $\pm$ standard deviation and Wilcoxon Rank-Sum Test (reference: RIcPSO) for RIcPSO against RecPSO and cPSO on CEC2014~\cite{cec2014} in $10$D.}\label{tab:cPSO10D}
\centering
\begin{tabular}{l|r@{$\,\pm\,$}l|c|r@{$\,\pm\,$}l|c|r@{$\,\pm\,$}l|c}
\hline\hline
              &      \multicolumn{3}{c|}{RIcPSO}    &   \multicolumn{3}{c|}{RecPSO}&   \multicolumn{3}{c}{cPSO}\\
\hline
$f_{1}$ & \multicolumn{3}{c|}{$2.78e+06\,\pm\,1.48e+06$} & $2.92e+06$ & $1.22e+06$ & = & $\mathbf{2.36e+06}$ & $\mathbf{1.72e+06}$ & =\\
$f_{2}$ & \multicolumn{3}{c|}{$4.08e+07\,\pm\,1.69e+07$} & $3.79e+07$ & $1.41e+07$ & = & $\mathbf{3.10e+07}$ & $\mathbf{1.34e+07}$ & -\\
$f_{3}$ & \multicolumn{3}{c|}{$7.23e+03\,\pm\,2.87e+03$} & $6.87e+03$ & $2.51e+03$ & = & $\mathbf{5.67e+03}$ & $\mathbf{3.22e+03}$ & -\\
$f_{4}$ & \multicolumn{3}{c|}{$4.44e+01\,\pm\,1.45e+01$} & $4.60e+01$ & $1.68e+01$ & = & $\mathbf{4.03e+01}$ & $\mathbf{1.71e+01}$ & =\\
$f_{5}$ & \multicolumn{3}{c|}{$\mathbf{2.02e+01}\,\pm\,\mathbf{1.38e+00}$} & $2.02e+01$ & $1.07e+00$ & = & $2.02e+01$ & $1.27e+00$ & =\\
$f_{6}$ & \multicolumn{3}{c|}{$3.45e+00\,\pm\,9.73e-01$} & $\mathbf{3.37e+00}$ & $\mathbf{7.43e-01}$ & = & $3.42e+00$ & $1.40e+00$ & =\\
$f_{7}$ & \multicolumn{3}{c|}{$1.84e+00\,\pm\,2.98e-01$} & $1.83e+00$ & $2.73e-01$ & = & $\mathbf{1.65e+00}$ & $\mathbf{3.35e-01}$ & -\\
\hline\hline
\end{tabular}
\end{table}

\begin{table}[p]
\caption{Average error $\pm$ standard deviation and Wilcoxon Rank-Sum Test (reference: RIcPSO) for RIcPSO against RecPSO and cPSO on CEC2014~\cite{cec2014} in $50$D.}\label{tab:cPSO50D}
\centering
\begin{tabular}{l|r@{$\,\pm\,$}l|c|r@{$\,\pm\,$}l|c|r@{$\,\pm\,$}l|c}
\hline\hline
              &      \multicolumn{3}{c|}{RIcPSO}    &   \multicolumn{3}{c|}{RecPSO}&   \multicolumn{3}{c}{cPSO}\\
\hline
$f_{1}$ & \multicolumn{3}{c|}{$4.37e+08\,\pm\,1.40e+08$} & $\mathbf{4.17e+08}$ & $\mathbf{1.08e+08}$ & = & $4.56e+08$ & $1.95e+08$ & =\\
$f_{2}$ & \multicolumn{3}{c|}{$4.53e+10\,\pm\,4.65e+09$} & $4.33e+10$ & $3.93e+09$ & - & $\mathbf{3.89e+10}$ & $\mathbf{6.45e+09}$ & -\\
$f_{3}$ & \multicolumn{3}{c|}{$1.27e+05\,\pm\,3.39e+04$} & $\mathbf{1.25e+05}$ & $\mathbf{3.15e+04}$ & = & $1.27e+05$ & $3.02e+04$ & =\\
$f_{4}$ & \multicolumn{3}{c|}{$7.13e+03\,\pm\,1.40e+03$} & $6.85e+03$ & $1.34e+03$ & = & $\mathbf{6.56e+03}$ & $\mathbf{1.63e+03}$ & =\\
$f_{5}$ & \multicolumn{3}{c|}{$\mathbf{2.11e+01}\,\pm\,\mathbf{4.49e-02}$} & $2.12e+01$ & $4.30e-02$ & = & $2.12e+01$ & $3.31e-02$ & =\\
$f_{6}$ & \multicolumn{3}{c|}{$6.11e+01\,\pm\,2.36e+00$} & $\mathbf{6.05e+01}$ & $\mathbf{2.33e+00}$ & = & $6.12e+01$ & $3.22e+00$ & =\\
$f_{7}$ & \multicolumn{3}{c|}{$4.17e+02\,\pm\,4.23e+01$} & $4.26e+02$ & $4.29e+01$ & = & $\mathbf{3.75e+02}$ & $\mathbf{5.81e+01}$ & -\\
\hline\hline
\end{tabular}
\end{table}

\begin{table}[p]
\caption{Average error $\pm$ standard deviation and Wilcoxon Rank-Sum Test (reference: RIcPSO) for RIcPSO against RecPSO and cPSO on CEC2014~\cite{cec2014} in $100$D.}\label{tab:cPSO100D}
\centering
\begin{tabular}{l|r@{$\,\pm\,$}l|c|r@{$\,\pm\,$}l|c|r@{$\,\pm\,$}l|c}
\hline\hline
              &      \multicolumn{3}{c|}{RIcPSO}    &   \multicolumn{3}{c|}{RecPSO}&   \multicolumn{3}{c}{cPSO}\\
\hline
$f_{1}$ & \multicolumn{3}{c|}{$2.29e+09\,\pm\,3.75e+08$} & $\mathbf{2.27e+09}$ & $\mathbf{3.88e+08}$ & = & $2.53e+09$ & $6.99e+08$ & =\\
$f_{2}$ & \multicolumn{3}{c|}{$1.87e+11\,\pm\,1.29e+10$} & $1.83e+11$ & $1.26e+10$ & = & $\mathbf{1.80e+11}$ & $\mathbf{1.31e+10}$ & -\\
$f_{3}$ & \multicolumn{3}{c|}{$3.76e+05\,\pm\,3.73e+04$} & $3.74e+05$ & $3.34e+04$ & = & $\mathbf{3.67e+05}$ & $\mathbf{4.10e+04}$ & =\\
$f_{4}$ & \multicolumn{3}{c|}{$3.68e+04\,\pm\,4.43e+03$} & $3.75e+04$ & $4.45e+03$ & = & $\mathbf{3.36e+04}$ & $\mathbf{4.14e+03}$ & -\\
$f_{5}$ & \multicolumn{3}{c|}{$2.13e+01\,\pm\,2.27e-02$} & $2.13e+01$ & $1.80e-02$ & = & $\mathbf{2.13e+01}$ & $\mathbf{1.95e-02}$ & =\\
$f_{6}$ & \multicolumn{3}{c|}{$\mathbf{1.47e+02}\,\pm\,\mathbf{3.31e+00}$} & $1.47e+02$ & $2.52e+00$ & = & $1.48e+02$ & $5.77e+00$ & =\\
$f_{7}$ & \multicolumn{3}{c|}{$1.72e+03\,\pm\,1.37e+02$} & $1.70e+03$ & $1.64e+02$ & = & $\mathbf{1.68e+03}$ & $\mathbf{1.38e+02}$ & =\\
\hline\hline
\end{tabular}
\end{table}


\begin{table}[p]
\caption{Average error $\pm$ standard deviation and Wilcoxon Rank-Sum Test (reference: RIcBFO) for RIcBFO against RecBFO and cBFO on CEC2014~\cite{cec2014} in $10$D.}\label{tab:cBFO10D}
\centering
\begin{tabular}{l|r@{$\,\pm\,$}l|c|r@{$\,\pm\,$}l|c|r@{$\,\pm\,$}l|c}
\hline\hline
              &      \multicolumn{3}{c|}{RIcBFO}    &   \multicolumn{3}{c|}{RecBFO}&   \multicolumn{3}{c}{cBFO}\\
\hline
$f_{1}$ & \multicolumn{3}{c|}{$8.83e+05\,\pm\,3.54e+05$} & $9.53e+05$ & $4.25e+05$ & = & $\mathbf{3.08e+05}$ & $\mathbf{2.02e+05}$ & -\\
$f_{2}$ & \multicolumn{3}{c|}{$6.23e+05\,\pm\,3.39e+05$} & $5.90e+05$ & $3.15e+05$ & = & $\mathbf{5.17e+04}$ & $\mathbf{2.49e+04}$ & -\\
$f_{3}$ & \multicolumn{3}{c|}{$1.14e+03\,\pm\,3.62e+02$} & $1.08e+03$ & $5.57e+02$ & = & $\mathbf{6.05e+01}$ & $\mathbf{3.60e+01}$ & -\\
$f_{4}$ & \multicolumn{3}{c|}{$\mathbf{1.99e+01}\,\pm\,\mathbf{1.54e+01}$} & $2.19e+01$ & $1.54e+01$ & = & $3.10e+01$ & $1.16e+01$ & =\\
$f_{5}$ & \multicolumn{3}{c|}{$2.03e+01\,\pm\,7.69e-02$} & $2.03e+01$ & $8.80e-02$ & = & $\mathbf{2.02e+01}$ & $\mathbf{8.63e-02}$ & -\\
$f_{6}$ & \multicolumn{3}{c|}{$\mathbf{5.88e+00}\,\pm\,\mathbf{6.44e-01}$} & $6.09e+00$ & $4.55e-01$ & = & $6.22e+00$ & $8.30e-01$ & =\\
$f_{7}$ & \multicolumn{3}{c|}{$7.24e-01\,\pm\,1.16e-01$} & $6.54e-01$ & $1.12e-01$ & - & $\mathbf{4.58e-01}$ & $\mathbf{1.52e-01}$ & -\\
\hline\hline
\end{tabular}
\end{table}

\begin{table}[p]
\caption{Average error $\pm$ standard deviation and Wilcoxon Rank-Sum Test (reference: RIcBFO) for RIcBFO against RecBFO and cBFO on CEC2014~\cite{cec2014} in $50$D.}\label{tab:cBFO50D}
\centering
\begin{tabular}{l|r@{$\,\pm\,$}l|c|r@{$\,\pm\,$}l|c|r@{$\,\pm\,$}l|c}
\hline\hline
              &      \multicolumn{3}{c|}{RIcBFO}    &   \multicolumn{3}{c|}{RecBFO}&   \multicolumn{3}{c}{cBFO}\\
\hline
$f_{1}$ & \multicolumn{3}{c|}{$1.04e+07\,\pm\,1.83e+06$} & $1.04e+07$ & $1.58e+06$ & = & $\mathbf{4.54e+06}$ & $\mathbf{1.22e+06}$ & -\\
$f_{2}$ & \multicolumn{3}{c|}{$6.20e+06\,\pm\,1.01e+06$} & $5.87e+06$ & $9.14e+05$ & = & $\mathbf{8.19e+05}$ & $\mathbf{2.76e+05}$ & -\\
$f_{3}$ & \multicolumn{3}{c|}{$9.09e+03\,\pm\,3.98e+03$} & $9.90e+03$ & $3.61e+03$ & = & $\mathbf{2.84e+03}$ & $\mathbf{6.59e+03}$ & -\\
$f_{4}$ & \multicolumn{3}{c|}{$\mathbf{1.91e+02}\,\pm\,\mathbf{3.12e+01}$} & $2.00e+02$ & $2.83e+01$ & = & $2.27e+02$ & $3.92e+01$ & +\\
$f_{5}$ & \multicolumn{3}{c|}{$2.07e+01\,\pm\,8.04e-02$} & $2.07e+01$ & $7.00e-02$ & = & $\mathbf{2.06e+01}$ & $\mathbf{9.20e-02}$ & =\\
$f_{6}$ & \multicolumn{3}{c|}{$6.11e+01\,\pm\,2.29e+00$} & $\mathbf{6.05e+01}$ & $\mathbf{2.69e+00}$ & = & $6.34e+01$ & $3.86e+00$ & +\\
$f_{7}$ & \multicolumn{3}{c|}{$8.95e-01\,\pm\,3.81e-02$} & $9.10e-01$ & $4.18e-02$ & = & $\mathbf{8.31e-01}$ & $\mathbf{6.77e-02}$ & -\\
\hline\hline
\end{tabular}
\end{table}

\begin{table}[p]
\caption{Average error $\pm$ standard deviation and Wilcoxon Rank-Sum Test (reference: RIcBFO) for RIcBFO against RecBFO and cBFO on CEC2014~\cite{cec2014} in $100$D.}\label{tab:cBFO100D}
\centering
\begin{tabular}{l|r@{$\,\pm\,$}l|c|r@{$\,\pm\,$}l|c|r@{$\,\pm\,$}l|c}
\hline\hline
              &      \multicolumn{3}{c|}{RIcBFO}    &   \multicolumn{3}{c|}{RecBFO}&   \multicolumn{3}{c}{cBFO}\\
\hline
$f_{1}$ & \multicolumn{3}{c|}{$6.81e+07\,\pm\,6.95e+06$} & $6.86e+07$ & $6.77e+06$ & = & $\mathbf{6.46e+07}$ & $\mathbf{1.54e+07}$ & =\\
$f_{2}$ & \multicolumn{3}{c|}{$8.72e+06\,\pm\,9.86e+05$} & $8.59e+06$ & $8.68e+05$ & = & $\mathbf{3.13e+06}$ & $\mathbf{5.78e+05}$ & -\\
$f_{3}$ & \multicolumn{3}{c|}{$1.07e+05\,\pm\,1.25e+04$} & $1.06e+05$ & $1.32e+04$ & = & $\mathbf{2.22e+04}$ & $\mathbf{1.53e+04}$ & -\\
$f_{4}$ & \multicolumn{3}{c|}{$\mathbf{4.63e+02}\,\pm\,\mathbf{4.63e+01}$} & $4.69e+02$ & $3.85e+01$ & = & $5.06e+02$ & $4.08e+01$ & +\\
$f_{5}$ & \multicolumn{3}{c|}{$2.09e+01\,\pm\,4.10e-02$} & $\mathbf{2.09e+01}$ & $\mathbf{4.37e-02}$ & = & $2.09e+01$ & $4.80e-02$ & =\\
$f_{6}$ & \multicolumn{3}{c|}{$1.42e+02\,\pm\,3.17e+00$} & $\mathbf{1.42e+02}$ & $\mathbf{3.47e+00}$ & = & $1.46e+02$ & $4.77e+00$ & +\\
$f_{7}$ & \multicolumn{3}{c|}{$1.42e+00\,\pm\,6.12e-02$} & $1.41e+00$ & $5.90e-02$ & = & $\mathbf{1.41e+00}$ & $\mathbf{2.35e-01}$ & =\\
\hline\hline
\end{tabular}
\end{table}


\begin{table}[p]
\caption{Average error $\pm$ standard deviation for Random Walk (RW) on CEC2014~\cite{cec2014} in $10$, $50$ and $100$D.}\label{tab:RW}
\centering
 \begin{tabular}{l|r@{$\,\pm\,$}l|c|r@{$\,\pm\,$}l|r@{$\,\pm\,$}lc}
\hline\hline
\multicolumn{8}{c}{Random Walk}\\
\hline
              &      \multicolumn{3}{c|}{$10$ dimensions}    &   \multicolumn{2}{c|}{$50$ dimensions}&   \multicolumn{2}{c}{$100$ dimensions}\\
\hline
$f_{1}$ & \multicolumn{3}{c|}{${2.59e+07}\,\pm\,{9.06e+06}$} & $2.65e+09$ & $3.44e+08$ &  $8.47e+09$ & $8.78e+08$ \\
$f_{2}$ & \multicolumn{3}{c|}{${3.10e+09}\,\pm\,{7.85e+08}$} & $1.60e+11$ & $1.20e+10$ &  $4.38e+11$ & $1.72e+10$ \\
$f_{3}$ & \multicolumn{3}{c|}{${2.05e+04}\,\pm\,{4.45e+03}$} & $2.24e+05$ & $1.94e+04$ &  $5.53e+05$ & $3.12e+04$ \\
$f_{4}$ & \multicolumn{3}{c|}{${2.86e+02}\,\pm\,{6.97e+01}$} & $4.15e+04$ & $5.11e+03$ &  $1.35e+05$ & $1.09e+04$ \\
$f_{5}$ & \multicolumn{3}{c|}{${2.04e+01}\,\pm\,{7.03e-02}$} & $2.12e+01$ & $3.96e-02$ &  $2.13e+01$ & $1.91e-02$ \\
$f_{6}$ & \multicolumn{3}{c|}{${9.32e+00}\,\pm\,{5.64e-01}$} & $7.25e+01$ & $1.38e+00$ &  $1.60e+02$ & $1.74e+00$ \\
\hline\hline
\end{tabular}
\end{table}


\end{document}